\documentclass[10pt,twocolumn,letterpaper]{article}

\usepackage{cvpr}              

\usepackage{graphicx}
\usepackage{amsmath}
\usepackage{amssymb}
\usepackage{booktabs}
\usepackage[accsupp]{axessibility}
\usepackage[table,xcdraw]{xcolor}
\usepackage{numprint} 
\usepackage{dirtytalk}
\usepackage{color,soul} 
\usepackage[shortlabels]{enumitem} 
\MakeRobust{\say} 

\usepackage[pagebackref,breaklinks,colorlinks]{hyperref}

\usepackage[capitalize]{cleveref}
\crefname{section}{Section}{Sections}
\Crefname{section}{Section}{Sections}
\Crefname{table}{Table}{Tables}
\crefname{table}{Tab.}{Tabs.}
\Crefname{figure}{Figure}{Figures}
\crefname{figure}{Figure}{Figures}
\Crefname{equation}{Equation}{Equations}
\crefname{equation}{Equation}{Equations}

\usepackage{multirow}
\usepackage{makecell}

\newcommand{\BY}{\kern0.0556em×\kern0.0556em}
\newcommand{\VEC}[1]{\ensuremath\mathbf{#1}}
\newcommand{\MAT}[1]{\ensuremath\mathbf{#1}}

\def\cvprPaperID{6} 
\def\confName{CVPR}
\def\confYear{2023}

\begin{document}

\title{Uncovering the Inner Workings of STEGO for \\Safe Unsupervised Semantic Segmentation}


\author{
Alexander Koenig 
\qquad
Maximilian Schambach
\qquad
Johannes Otterbach
\\
\\ 
Merantix Momentum\\
\tt\small \{firstname.lastname\}@merantix.com
}
\maketitle

\begin{abstract}
    Self-supervised pre-training strategies have recently shown impressive results for training general-purpose feature extraction backbones in computer vision. In combination with the Vision Transformer architecture, the DINO self-distillation technique has interesting emerging properties, such as unsupervised clustering in the latent space and semantic correspondences of the produced features without using explicit human-annotated labels. 
    The STEGO method for unsupervised semantic segmentation contrastively distills feature correspondences of a DINO-pre-trained Vision Transformer and recently set a new state of the art. However, the detailed workings of STEGO have yet to be disentangled, preventing its usage in safety-critical applications.
    
    This paper provides a deeper understanding of the STEGO architecture and training strategy by conducting studies that uncover the working mechanisms behind STEGO, reproduce and extend its experimental validation, and investigate the ability of STEGO to transfer to different datasets. Results demonstrate that the STEGO architecture can be interpreted as a semantics-preserving dimensionality reduction technique.
\end{abstract}
\vspace{-3mm}

\section{Introduction}
\label{sec:intro}
Semantic segmentation is the task of assigning pixel-wise class labels to an image. Its applications range from scene understanding for autonomous navigation~\cite{setr,ganav}, environmental monitoring~\cite{agri,jordan}, to medical imaging~\cite{unet, her2net} to name a few. Current state-of-the-art deep learning-based semantic segmentation approaches are trained in a supervised fashion and thus require a large amount of training data, \ie, explicit pixel-wise target labels, also known as segmentation maps. However, freely available open-source datasets with dense segmentation labels for industrial applications are scarce and costly to obtain as opposed to per-image class labels such as those used in popular image datasets like ImageNet~\cite{imagenet2015}, CIFAR~\cite{cifar}, or MNIST~\cite{mnist}. 

To alleviate this shortcoming, researchers recently turned to self-supervised learning methods to retrieve semantic information from images without using human-annotated labels. The field recently gained traction with self-supervised representation learning algorithms such as \mbox{SimCLR}~\cite{chen20simclr}, {BYOL}~\cite{byol}, and {SwAV}~\cite{swav}, with which feature extraction backbones can be learned without the explicit use of target labels. The image representations generated by these backbones can then be used for various downstream tasks. Among those algorithms, the DINO pre-training strategy~\cite{caron2021emerging} is particularly notable. Combined with the recent Visual Transformer (ViT) architecture~\cite{dosovitskiy2021an}, the DINO pre-training objective produces semantically consistent patch-wise image embeddings that can be effectively used for various downstream tasks such as image classification~\cite{caron2021emerging}, part co-segmentation~\cite{amir2021deep}, video instance segmentation~\cite{caron2021emerging}, or neural rendering~\cite{tschernezki22neural}. In the remainder, we refer to a ViT trained using the DINO pre-training strategy as the DINO backbone, or simply, DINO.
These feature extraction backbones also facilitate self-supervised semantic segmentation. For instance, the STEGO architecture and training strategy~\cite{hamilton2022unsupervised} builds on DINO and post-processes its expressive features through a self-supervised correlation loss, which distills feature correspondences. Despite being compute-efficient by freezing the DINO backbone, the STEGO approach demonstrates impressive capabilities on unsupervised semantic segmentation benchmarks.

Given the promises of these methods, it is essential to understand their transferability to real-world settings before deployment to ensure reproducibility, robustness, and safety. However, looking closely at STEGO, we struggled to reproduce some of the results reported in the original publication~\cite{hamilton2022unsupervised}. Additionally, self-supervised learning mechanisms are intricate and still poorly understood. In particular, experimental evaluation has yet to clearly separate contributions in ablation studies and benchmark performance across multiple datasets to ensure robust performance in safety-critical applications. Concerning self-supervised segmentation using STEGO, we identified some missing benchmark results and ablation studies investigating the contribution of this new approach in more detail.
%

In this paper, we provide a deeper understanding of the STEGO architecture and training strategy to ensure its well-informed usage and motivate further development. Particularly, our contributions are as follows.

\begin{enumerate}
    \item We reproduce and extend STEGO's experimental validation, demonstrating a stronger baseline performance of the DINO backbone than reported in the STEGO paper~\cite{hamilton2022unsupervised} and reporting missing evaluation metrics for a more comprehensive evaluation.
    \item We disentangle STEGO's working mechanisms and attribute its superior unsupervised performance to both dimensionality reduction of the backbone's embeddings and a non-linear projection that yields more semantically clustered representations.
    \item We compare STEGO with well-established dimensionality reduction techniques across different dimensions and datasets, revealing STEGO's better suitability for unsupervised clustering and providing insights into its sensitivity to the target dimension.


\end{enumerate}


\section{Background}
\label{sec:background}

\subsection{Related Works}
Methods addressing label scarcity in deep semantic segmentation can broadly be taxonomized into \emph{weakly supervised}, and \emph{unsupervised} methods \cite{surveyLabelEfficient}.
Weakly supervised algorithms for semantic segmentation often rely on coarse supervision, \eg, by training a segmentation model only through image-level annotations \cite{Ahn_2018_CVPR, Wang_2020_CVPR} or bounding box labels \cite{Oh_2021_CVPR}. Other weakly supervised approaches use incompletely labeled segmentation datasets, also known as semi-supervision \cite{Lai_2021_CVPR}. 

On the other hand, unsupervised semantic segmentation algorithms use no explicit labels. A core idea behind unsupervised semantic segmentation is clustering dense visual descriptors of an image into perceptual groups. Some algorithms frame this task as a graph partitioning problem, either on the input image \cite{normcuts} or, more recently, on deep ViT-features \cite{Melas-Kyriazi_2022_CVPR}. Furthermore, standard $k$-means-style clustering \cite{amir2021deep}, or matrix factorization \cite{Collins_2018_ECCV} of features from pre-trained backbones were successfully used to detect related ontologies for unsupervised semantic segmentation. Other methods use self-supervision to learn visual descriptors for unsupervised segmentation. For instance, PiCIE \cite{Cho2021PiCIE} clusters pixel-level features by optimizing a self-supervised loss that enforces invariance to photometric and equivariance to geometric transformations. Self-supervision is also often used for feature distillation of pre-trained backbones before clustering. For example, the MaskContrast~\cite{MaskContrast} algorithm uses a pre-trained network to extract a binary mask of an image's main object, also referred to as saliency estimation, contrastively trains a separate network that maximizes agreement of the pixel embeddings with the extracted object mask, and finally clusters the pixel embeddings using $k$-means. MaskDistill~\cite{MaskDistill} uses object masks from clustered ViT-features as pseudo-ground-truth labels to train a second network on high-confidence segmentation masks. While previous feature distillation approaches introduce a proxy network, the STEGO~\cite{hamilton2022unsupervised} algorithm directly utilizes a ViT backbone, distills the produced latent features, and clusters them into distinct ontologies -- promising a simple and efficient training strategy.

\subsection{STEGO Architecture}

\begin{figure*}[ht]
    \centering
    \includegraphics[width=\linewidth]{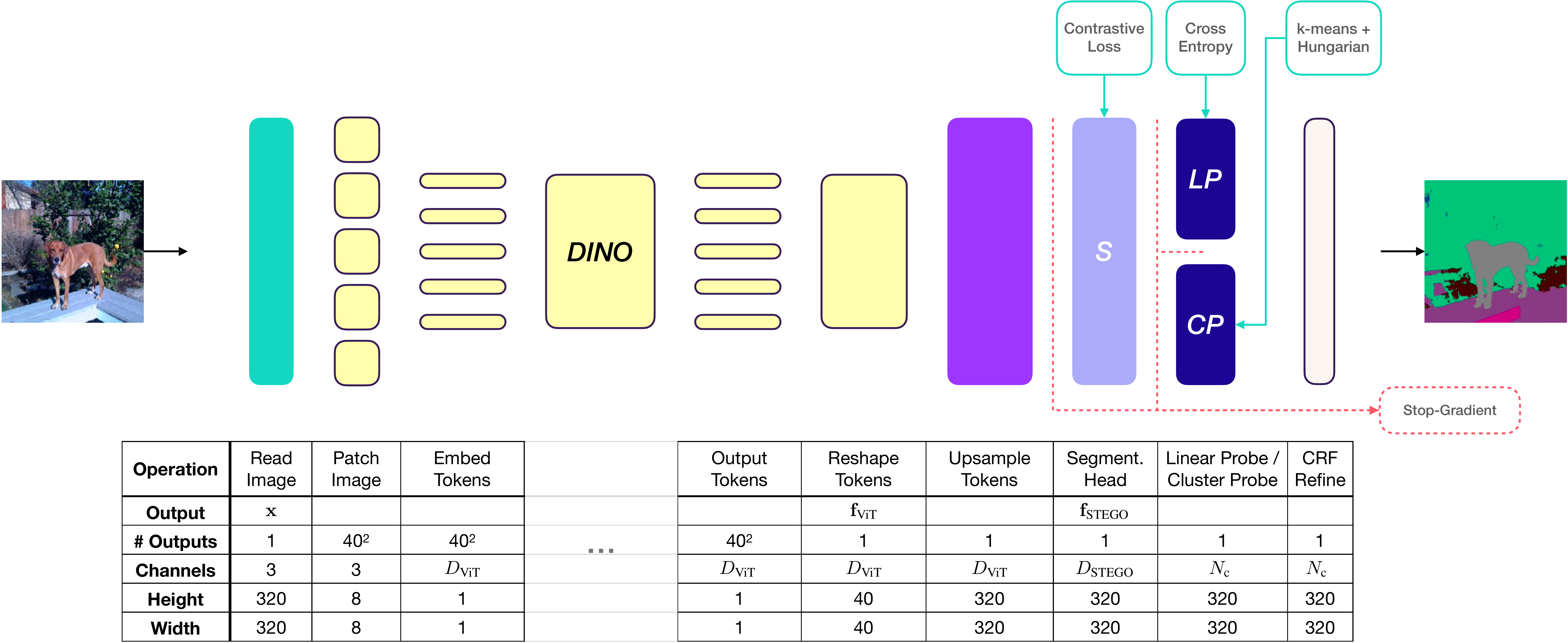}
    \caption{Architecture of the STEGO validation pipeline. The top of the figure shows abstractions of the operations and their training losses. The table indicates each operation's outputs and their dimensions. $\VEC{x}$ is the input image, \emph{DINO} is the frozen ViT backbone, $\VEC{f}_\textrm{ViT}$ is an image-like feature map (\ie, reshaped ViT tokens), $S$ is the segmentation head, and the Linear Probe \emph{(LP)} and Cluster Probe \emph{(CP)} evaluate the segmentation head output $\VEC{f}_\textrm{STEGO}$ and map to $N_\textrm{c}$ class probabilities. The stop-gradient module ensures that the optimizations of $S$, \emph{LP}, and \emph{CP} do not influence each other and that label information can neither propagate into $S$ nor \emph{DINO}. We display the validation pipeline, which uses 320\BY 320 images. The training pipeline processes 224\BY 224 images, does not upsample the ViT feature map $\VEC{f}_\textrm{ViT}$, and does not use CRF by default.}
    \label{fig:architecture}
\end{figure*}

The architecture of the STEGO pipeline is given in \cref{fig:architecture}. The approach uses an ImageNet-pre-trained DINO backbone~\cite{caron2021emerging} for feature extraction. Before being fed into the ViT, an input image $\VEC{x}$ is patched into 8\BY 8 image patches which are subsequently mapped into a latent representation using a simple linear layer. A learnable positional encoding is then added to these embedded patches, also called tokens. Multiple Transformer layers then process these tokens. Finally, the ViT's output tokens are reshaped into a $D_\textrm{ViT}$-dimensional image-like feature map $\VEC{f}_\textrm{ViT}$. The STEGO architecture adds three main modules on top of DINO. First, an optional bilinear upsampling layer is added to upsample the feature map by a factor of 8 to regain the original input image resolution. Then, an unsupervised segmentation head projects the DINO features from $D_\textrm{ViT}$ to $D_\textrm{STEGO}$ dimensions. Here, the dimensionality $D_\textrm{ViT}$ of the ViT features depends on the choice of the backbone ($D_\textrm{ViT}=384$ for ViT-Small, $768$ for ViT-Base) and the target dimension $D_\textrm{STEGO}$ is a hyper-parameter. Lastly, a cluster or linear probe is trained with few labels to assign each pixel a probability of each of the $N_\textrm{C}$ semantic target classes. The probes evaluate the feature quality of the segmentation head's output. A stop-gradient operation ensures that the simultaneous optimizations of both probes and the segmentation head do not influence each other and that supervised label information propagates neither into the segmentation head nor the backbone. Finally, the output is optionally refined using a Conditional Random Field (CRF)~\cite{crf}.

\begin{figure}[ht]
    \centering
    \includegraphics[width=\linewidth]{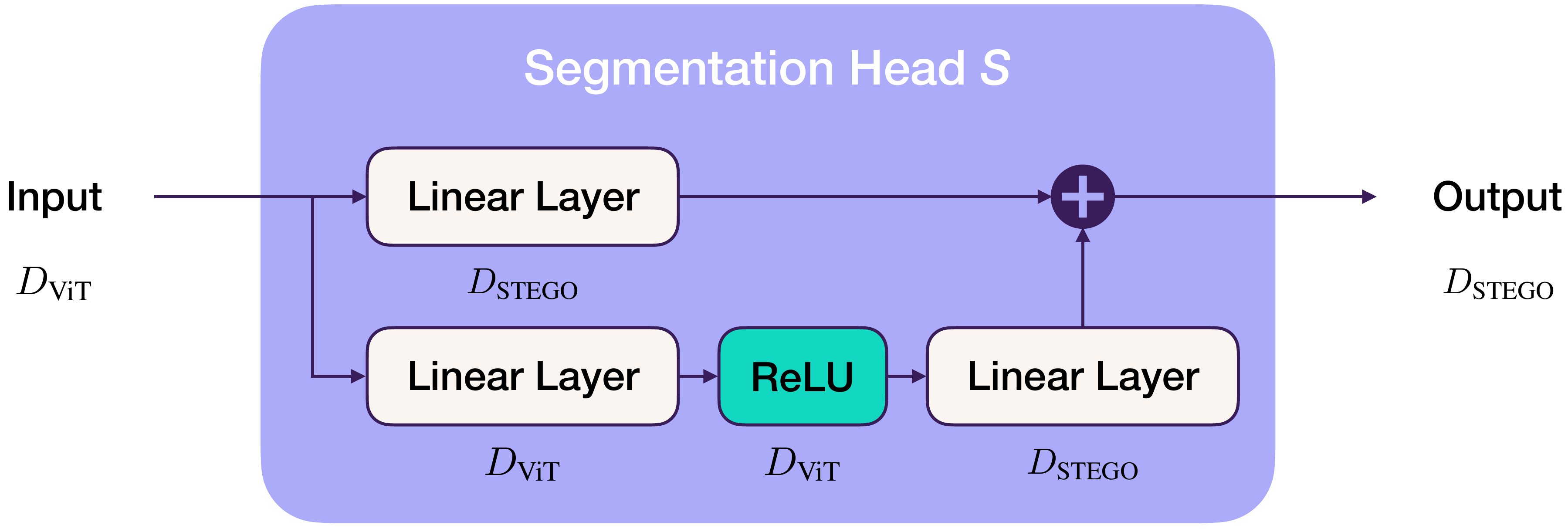}
    \caption{Architecture of the non-linear segmentation head $S$. The module projects the dimension of a vector from $D_\textrm{ViT}$ to $D_\textrm{STEGO}$.}
    \label{fig:seghead}
\end{figure}

As shown in \cref{fig:seghead}, the STEGO segmentation head consists of several linear layers and a non-linear ReLU~\cite{relu} activation function. The segmentation head processes each feature separately by applying a 1\BY 1 convolution to the feature map. That is, neighboring features do not influence each other in this head. Moreover, the module performs a projection of the $D_\textrm{ViT}$-dimensional input vector into a $D_\textrm{STEGO}$-dimensional output embedding space. The authors propose to reduce the channel dimension in the segmentation head, \ie, $D_\textrm{STEGO} < D_\textrm{ViT}$, and specify $D_\textrm{STEGO}=70$.

\subsection{STEGO Training Strategy}


Hamilton \etal's~\cite{hamilton2022unsupervised} contrastive framework is based on ViT token pairs as opposed to image pairs which are often used in the context of contrastive self-supervised training strategies in computer vision~\cite{chen20simclr,caron2021emerging}. Hamilton \etal define token pairs using the cosine similarity of two ViT tokens: A pair is considered positive when the cosine similarity of two tokens exceeds a certain threshold; otherwise, they are considered negative.
Their \emph{contrastive correlation loss} is used to train the segmentation head and push the features of positive token pairs towards alignment, corresponding to a cosine similarity of 1, while negative pairs are pushed towards a cosine similarity of --1.

The loss is implemented as follows. First, they assemble a pair of images $(\VEC{x}, \VEC{x}^\prime)$ and obtain the corresponding DINO feature maps $(\VEC{f}_\textrm{ViT}, \VEC{f}_\textrm{ViT}^\prime)$ as well as segmentation head outputs $(\VEC{f}_\textrm{STEGO}, \VEC{f}_\textrm{STEGO}^\prime)$. Subsequently, they calculate the cosine similarities between all token pairs $\VEC{f}_\textrm{ViT}$ and $\VEC{f}_\textrm{ViT}^\prime$ to obtain a feature correspondence tensor $\MAT{C}_\text{ViT}$. 
Each entry $\MAT{C}_{\textrm{ViT}, hwij} \in [-1,1]$ in the feature correspondence tensor is the cosine similarity between a token in $\VEC{f}_\textrm{ViT}$ and a token in $\VEC{f}_\textrm{ViT}^\prime$ where $(h,w)$ and $(i,j)$ represent the spatial locations of the tokens in the two feature maps, respectively. 
Analogously, they calculate the correspondence tensor $\MAT{C}_\textrm{STEGO}$ of the segmentation head outputs $(\VEC{f}_\textrm{STEGO}, \VEC{f}_\textrm{STEGO}^\prime)$. Their contrastive correlation loss 
\begin{equation}
    L_\textrm{corr}(\VEC{x}, \VEC{x}', b) = - \sum\limits_{hwij} (\MAT{C}_{\textrm{ViT}, hwij} - b) \MAT{C}_{\textrm{STEGO}, hwij}
\end{equation}
 pushes an entry in the STEGO feature correspondence tensor $\MAT{C}_\textrm{STEGO}$ towards 1 if the respective entry in $\MAT{C}_\textrm{ViT}$ exceeds the threshold $b$ in the case of a positive token pair and towards --1 otherwise. The parameter $b$ can be interpreted as a form of negative pressure. The paper mentions several strategies to stabilize the loss and prevent collapse, such as spatial centering of the correspondence tensor $\MAT{C}_\textrm{ViT}$ and zero-clamping of the correspondence tensor $\MAT{C}_\textrm{STEGO}$.

The choice of image pairs $(\VEC{x}, \VEC{x}')$ contributes towards the amount of positive and negative learning signal from the loss function. In total, the loss uses three image pairs: the image with itself $(\VEC{x},\VEC{x})$, the image with one of its 7 nearest neighbors $(\VEC{x},\VEC{x}_\textrm{knn})$, and the image with a randomly chosen one $(\VEC{x}, \VEC{x}_\textrm{rand})$.
While the first two image pairs provide a positive training signal, since many token pairs have similar features, the latter one contributes mostly negative signals because most token pairs will likely be dissimilar, assuming the dataset is large and diverse.
Finally, the overall loss
\begin{equation}
    L = \sum_{i\in\{\textrm{self}, \textrm{knn}, \textrm{rand}\} } \lambda_{i} L_\textrm{corr}(\VEC{x}, \VEC{x}_i, b_i)
\end{equation}
is calculated as the weighted sum of the correlation losses of the three individual pairs. Overall, the STEGO loss introduces six hyperparameters, $\lambda_\textrm{self}$, $\lambda_\textrm{knn}$, $\lambda_\textrm{rand}$, $b_\textrm{self}$, $b_\textrm{knn}$, and $b_\textrm{rand}$, which require careful tuning.

\section{Reestablishing Baselines}
\label{sec:repro}
%
Hamilton \etal show that STEGO significantly outperforms other unsupervised segmentation algorithms such as PiCIE~\cite{Cho2021PiCIE} on the Cocostuff-27~\cite{Caesar_2018_CVPR} dataset. Notably, they demonstrate that STEGO outperforms the DINO baseline, where segmentation is performed on the DINO features directly. However, our preliminary experiments had indicated that the performance of the DINO backbone alone is already highly competitive, \ie the DINO backbone produces highly semantic features -- an observation in line with the original presentation~\cite{caron2021emerging}. This discrepancy motivated us to conduct a controlled reproducibility study of the result stated in the original STEGO paper~\cite{hamilton2022unsupervised}.

\subsection{Experimental Setup}
\label{sec:experimental-setup}
\paragraph{Evaluation, datasets, metrics.}
Recent works on self-supervised learning evaluate the expressivity and generalizability of features on several downstream tasks. Common evaluation protocols in computer vision involve training linear or $k$-NN classifiers using the ImageNet benchmark~\cite{chen20simclr, caron2021emerging} or semantic segmentation challenges~\cite{byol}. We adhere to the protocol in the original presentation to evaluate the features produced by the STEGO segmentation head. 

Hamilton \etal~\cite{hamilton2022unsupervised} evaluate the segmentation head features in two styles. First, they propose an \emph{unsupervised cluster probe}, \ie a cosine similarity-based $k$-means algorithm~\cite{MacQueen1967} that detects $N_\textrm{C}$ clusters in the segmentation head's representation space. The identified clusters are mapped to human-interpretable labels using a linear sum assignment solved via the Hungarian algorithm~\cite{kuhn1955hungarian} on the ground-truth labels. 
Second, they assess the general feature quality using a \emph{supervised linear probe}, \ie a linear layer processing the extracted fixed segmentation head outputs, which is trained via a standard cross-entropy loss with labels from the ground-truth segmentation map.

To stay consistent with the original work, we use the Cocostuff-27~\cite{Caesar_2018_CVPR}, Cityscapes~\cite{Cordts_2016_CVPR}, and Potsdam-3~\cite{Ji_2019_ICCV} datasets for all of our experiments. We report the validation mean Intersection over Union (mIoU) and accuracy as the primary evaluation metrics to be directly comparable to the original presentation. We follow the exact evaluation procedure from the STEGO repository by reporting results on the validation set and cherry-picking models performing best on the validation unsupervised mIoU benchmark.

\begin{table}[t]
\centering
\small
\begin{tabular}{lrrrr}
\toprule
\multirow{2}{*}{\vspace{-3mm}\thead{Parameter}} & \multicolumn{4}{c}{\thead{\textbf{Configuration}} } \\ \cmidrule(lr){2-5} 
                                        & \multicolumn{3}{c}{Actual} & Reported \\ 
\midrule
{Dataset} & Cocostuff & Cityscapes & Potsdam & * \\ 
{Train steps \hspace{-1mm}} & {7000} & {7000} & 5000 & - \\  
{Batch size} & {32} & {32} & 16 & 32 \\ 
{Crop type} & {5-crop} & {5-crop} & No crop & 5-crop \\ 
{Backbone} & {ViT-B} & {ViT-B} & ViT-S & ViT-B \\ 
{0-clamp} & {-} & {-} & True & True \\ 
{Pointwise} & {True} & {False} & True & True \\ 
$D_\textrm{STEGO}$ & {90} & {100} & 70 & 70 \\ 
$\lambda_\textrm{rand}$ & {0.15} & {0.91} & 0.63 & * \\ 
$\lambda_\textrm{knn}$ & {1.00} & {0.58} & 0.25 & * \\ 
$\lambda_\textrm{self}$ & {0.10} & {1.00} & 0.67 & * \\ 
$b_\textrm{rand}$ & {1.00} & {0.31} & 0.76 & * \\ 
$b_\textrm{knn}$ & {0.20} & {0.18} & 0.02 & * \\ 
$b_\textrm{self}$ & {0.12} & {0.46} & 0.08 & *  \\
\bottomrule
\end{tabular}
\vspace{-1mm}
\caption{Training configuration extracted from the pre-trained models of STEGO and configuration reported in the paper~\cite{hamilton2022unsupervised}. 
The first 5 parameters are generic training parameters that also apply to our \say{non-STEGO experiments} (\ie, our baseline experiments from \cref{sec:repro} and the experiments in \cref{sec:dimred}). The remaining parameters are STEGO-specific training parameters. 
We show only parameters that are different between the datasets or deviate from the original paper's information. For the remaining parameters, we refer readers to our supplementary material. Note that \say{-} means the information was unavailable in the code or the paper, and \say{*} means multiple results were reported.}
\label{tab:config}
\end{table}

\begin{table*}[ht]
\small
\centering
\begin{tabular}{lrrrrrrrrrrrr}
\toprule
\multirow{3}{*}{\vspace{-9mm}Method} & \multicolumn{4}{c}{\textbf{Cocostuff}} & \multicolumn{4}{c}{\textbf{Cityscapes}} & \multicolumn{4}{c}{\textbf{Potsdam}} \\ \cmidrule(lr){2-5}\cmidrule(lr){6-9}\cmidrule(lr){10-13}
  & \multicolumn{2}{c}{\thead{Unsupervised\\ Cluster probe}} & \multicolumn{2}{c}{\thead{Supervised\\ Linear probe}} 
  & \multicolumn{2}{c}{\thead{Unsupervised\\ Cluster probe}} & \multicolumn{2}{c}{\thead{Supervised\\ Linear probe}}
  & \multicolumn{2}{c}{\thead{Unsupervised\\ Cluster probe}} & \multicolumn{2}{c}{\thead{Supervised\\ Linear probe}} 
  \\ \cmidrule(lr){2-3}\cmidrule(lr){4-5}\cmidrule(lr){6-7}\cmidrule(lr){8-9}\cmidrule(lr){10-11}\cmidrule(lr){12-13}
  & \thead{Acc} & \thead{mIoU} & \thead{Acc} & \thead{mIoU} & \thead{Acc} & \thead{mIoU} & \thead{Acc} & \thead{mIoU} & \thead{Acc} & \thead{mIoU} & \thead{Acc} & \thead{mIoU} \\ \midrule
STEGO (theirs)& 56.9 & {28.2} & 76.1 & 41.0 & 73.2 & {21.0} & - & - & 77.0 & {-} & - & - \\
STEGO (ours)& {\color[HTML]{13D9C1} $^\checkmark$56.9} & {{\color[HTML]{13D9C1} $^\checkmark$28.2}} & {\color[HTML]{13D9C1} $^\checkmark$76.1} & {\color[HTML]{13D9C1} $^\checkmark$41.1} & {\color[HTML]{13D9C1} $^\checkmark$73.2} & {{\color[HTML]{13D9C1} $^\checkmark$21.0}} & {89.6} & {28.0} & {\color[HTML]{13D9C1} $^\checkmark$77.0} & {{62.6}} & {85.9} & {74.8} \\
DINO (theirs)& 30.5 & {9.6} & 66.8 & 29.4 & - & {-} & - & - & - & {-} & - & - \\
DINO (ours)& { \color[HTML]{9e36ff} $^\dagger$42.4} & {{\color[HTML]{9e36ff}  $^\dagger$13.0}} & {\color[HTML]{9e36ff}  $^\dagger$75.8} & {\color[HTML]{FF5565} $^\ddagger$44.4} & 52.6 & {15.2} & {\color[HTML]{FF5565} $^\ddagger$91.3} & {\color[HTML]{FF5565} $^\ddagger$34.9} & 71.3 & {54.3} & 84.5 & 72.8 \\ 
\bottomrule
\end{tabular}
\caption{Validation results of reproducibility study showing accuracy (\say{Acc}) and mIoU in \% for the different evaluation styles and datasets. Values reported in the original paper~\cite{hamilton2022unsupervised} are marked \say{theirs}, and missing ones are denoted with \say{-}. Our results evaluating the pre-trained models are marked \say{STEGO (ours)}. \say{DINO (ours)} is a re-training of the probes directly on the DINO backbone. Successful reproduction is denoted with {\color[HTML]{13D9C1} $\checkmark$} and a stronger DINO performance than reported in \cite{hamilton2022unsupervised} is indicated with {\color[HTML]{9e36ff} $\dagger$} and {\color[HTML]{FF5565} $\ddagger$}. All evaluations post-process with CRF.}
\label{tab:repro_results}
\end{table*}
\paragraph{Model configuration.}
\cref{tab:config}'s middle column shows the actual training configuration we retrieved from each dataset's pre-trained STEGO model shipped with the paper's source code\footnote{\tt \href{https://github.com/mhamilton723/STEGO}{github.com/mhamilton723/STEGO}}. The table's right-most column mentions the parameters given in their report. Notably, some information in the \emph{actual configuration} differs from the data reported in the original publication.

The data in \cref{tab:config} offers other interesting insights. The STEGO embedding dimension $D_\textrm{STEGO}$ is different for all datasets, and the dimension reported in the paper only matches the Potsdam dataset. The training configuration for the Potsdam dataset differs from the other ones, \ie using a smaller backbone, fewer training steps, no five-cropping of training images, and a smaller batch size. Furthermore, the loss hyperparameters are remarkably different for each dataset. We also report the hyperparameters for the Potsdam dataset here, which are not given in the original paper~\cite{hamilton2022unsupervised}.

We use the \say{actual configuration} retrieved from the pre-trained models without performing additional hyperparameter optimizations to remain comparable with Hamilton \etal's~\cite{hamilton2022unsupervised} results. Our source code is publicly available\footnote{\tt \href{https://github.com/merantix-momentum/stego-studies}{github.com/merantix-momentum/stego-studies}} and only applies minimal changes to the original code, \eg, for more extensive logging, which are clearly highlighted.

\subsection{Results and Discussion}
%
\cref{tab:repro_results} displays the results of our reproducibility study across all three benchmarking datasets. First, we show that using the pre-trained STEGO models provided by the authors, we are able to reproduce the results from the paper (highlighted with {\color[HTML]{13D9C1} $\checkmark$}), which validates our evaluation pipeline. Moreover, we paint a more comprehensive picture by reporting missing metrics for the Cityscapes and Potsdam datasets. 

Second, we directly evaluate the features produced by the DINO backbone by re-training the cluster and linear probe on the backbone while fixing all other training settings. We find that the DINO baseline results are consistently better than those provided in the paper~\cite{hamilton2022unsupervised} (highlighted with {\color[HTML]{9e36ff} $\dagger$}). This is surprising as we use the evaluation pipeline provided by the authors and do not introduce changes that could influence the training performance.

Notably, for linear probe-style evaluation, the DINO baseline performs approximately on par with the STEGO approach, sometimes even outperforming STEGO (highlighted with {\color[HTML]{FF5565} $\ddagger$}). 
This result means that the features produced by DINO and the STEGO segmentation head exhibit similar linear separability. DINO demonstrates a remarkable ability to provide semantically meaningful features and generalize to different datasets despite being pre-trained solely on ImageNet~\cite{imagenet2015}, while Hamilton \etal~\cite{hamilton2022unsupervised} fine-tuned the segmentation head on the respective datasets. DINO's performance is particularly noteworthy, given the significant domain shifts between ImageNet and Potsdam's aerial imagery and ego-camera footage from Cityscapes.

However, as shown in \cref{tab:repro_results}, the STEGO embeddings are better suited for the unsupervised cluster probe task than the raw DINO features. 
Hamilton \etal argue that \say{\textit{[d]ue to the feature distillation process, STEGO's segmentation features tend to form clear clusters}} (page 6, \cite{hamilton2022unsupervised}). While more \say{\textit{distinct clusters}} (page 1, \cite{hamilton2022unsupervised}) could explain STEGO's superior $k$-means clustering performance, there might be additional possible explanations.

First, STEGO can adapt to a domain shift to the new data distribution since it includes a trainable segmentation head. It needs to be clarified what the contribution of the STEGO loss for adapting to new datasets really is, as opposed to the DINO pre-training objective, which could also be used to fine-tune the backbone. For instance, can one fine-tune the ViT backbone -- either entirely, using adapters~\cite{houlsbyAdapters}, or additional linear heads -- using the DINO loss on a new dataset and obtain a performance competitive with STEGO? As Caron \etal~\cite{caron2021emerging} show, DINO provides semantically meaningful representations. The DINO loss is better understood due to its pervasion in the community and involves less hyperparameter tuning.

Second, STEGO's dimensionality reduction from $D_\textrm{ViT}$ to $D_\textrm{STEGO}$ (a factor of $ \approx 8$ for Cocostuff and Cityscapes) could in itself be a reason for a higher unsupervised clustering performance. The $k$-means algorithm used for clustering minimizes the sum of squared distances between each data point and its cluster centroid. Due to the curse of dimensionality, distances between data points in higher dimensions become exponentially larger and more uniform. This observation also holds for the cosine distance-based variant of $k$-means applied in the STEGO paper. Less significant distances make it harder to identify clusters in higher dimensional spaces. Hence, it could be that STEGO preserves the semantics of DINO features while projecting them into a lower-dimensional space where $k$-means performs better.
Again, the question remains whether a similar result could be obtained by training a feature downsampling layer or segmentation head using the DINO strategy.

The STEGO paper~\cite{hamilton2022unsupervised} does not explore these issues experimentally. 
Particularly, the research community has yet to disentangle the mechanisms behind STEGO's superior performance in the unsupervised clustering case with experimental results. In the remainder of this paper, we take a closer look at STEGO and test whether the method can be interpreted as a dimensionality-reduction technique that preserves DINO's semantic feature correspondences.

\section{Disentangling STEGO's Working Principles}
\label{sec:dimred}
%

We work towards disentangling the functional mechanisms behind STEGO by re-training the segmentation head with different output dimensions $D_\textrm{STEGO}$. Consequently, we draw conclusions from STEGO's performance on the linear and cluster probe downstream tasks. These ablations also give practitioners an intuition for how sensitive STEGO is towards its output dimension $D_\textrm{STEGO}$ and serve as another reproducibility experiment.
Previously, we argued that STEGO may be interpretable as a semantics-preserving dimensionality reduction technique. Hence, we compare STEGO with two standard methods that can reduce the dimensionality of the high-dimensional ViT features.

\subsection{Dimensionality Reduction Baselines}

\paragraph{Principal component analysis.} Principal component analysis (PCA)~\cite{pca} is a well-established unsupervised dimensionality reduction technique. First, PCA identifies a set of orthogonal vectors, known as principal components, that capture the most variance in the dataset. Secondly, the algorithm uses the top-$k$ principal components to linearly project the dataset into a lower-dimensional space.

\begin{figure}
    \centering
    \includegraphics[width=0.9\linewidth]{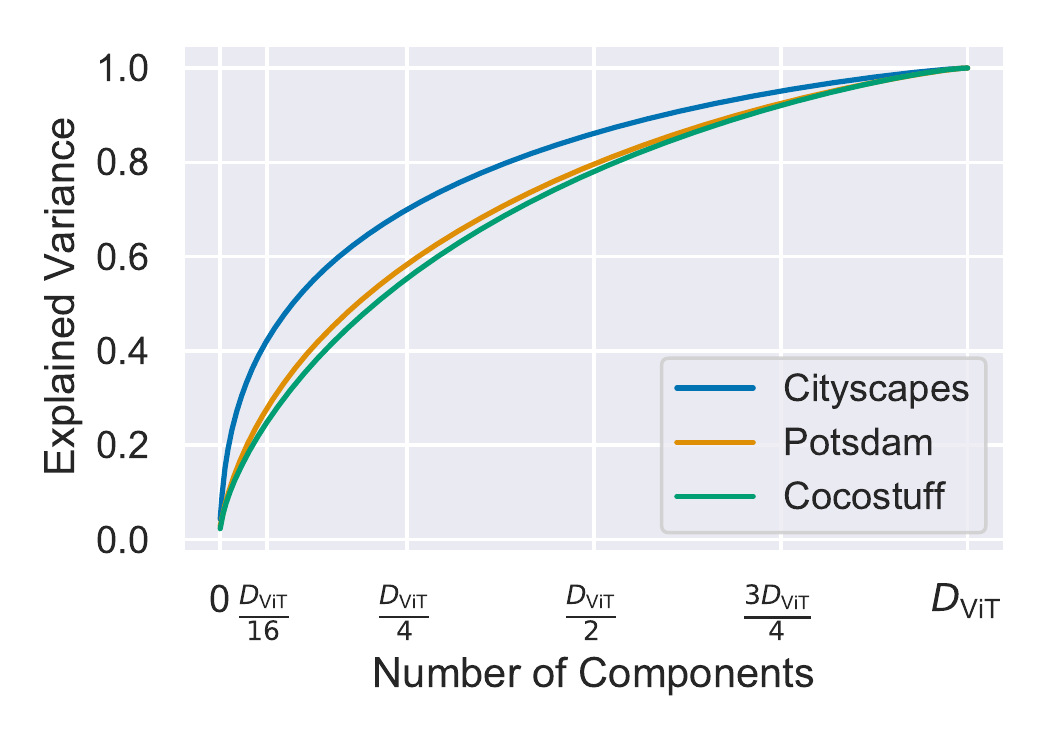}\vspace{-3mm}
    \caption{Cumulative explained variance of the principal components of DINO features for all datasets. We fit PCA on at most 5000 randomly selected training images, \ie, $\approx$\,3M ViT-tokens, to keep the memory consumption tractable. The $x$-axis is normalized as different embedding dimensions are used for the three datasets.}
    \label{fig:pca}
\end{figure}

\cref{fig:pca} shows the cumulative explained variance of the principal components of the DINO embeddings for all three investigated datasets. For Cityscapes, PCA can explain roughly 75\,\% of the dataset's variance using only roughly 33\,\% of the number of components. For the other datasets, more components are required to explain the same variance. The steep cumulative explained variance curve, especially for Cityscapes, indicates great potential for dimensionality reduction techniques operating on the ViT features. 

\vspace{-5mm}
\paragraph{Random Projection.} Another compute-efficient dimensionality reduction technique commonly used in the machine learning context is Random Projection (RP)~\cite{binghamRandomProj,DasguptaRandomProj}, which linearly transforms high-dimensional feature vectors into a randomly chosen subspace. We use Gaussian RP, which initializes the projection matrix with orthogonal column vectors drawn from a normal distribution. The algorithm approximately retains pair-wise distances with high probability~\cite{Lindenstrauss1984}, which makes RP well-suited for clustering with $k$-means~\cite{kMeanswRandomProj}.

\subsection{Results and Discussion}
\label{sec:discussion_dimensions}

\begin{figure*}[ht]
    \centering
    \includegraphics[width=\linewidth]{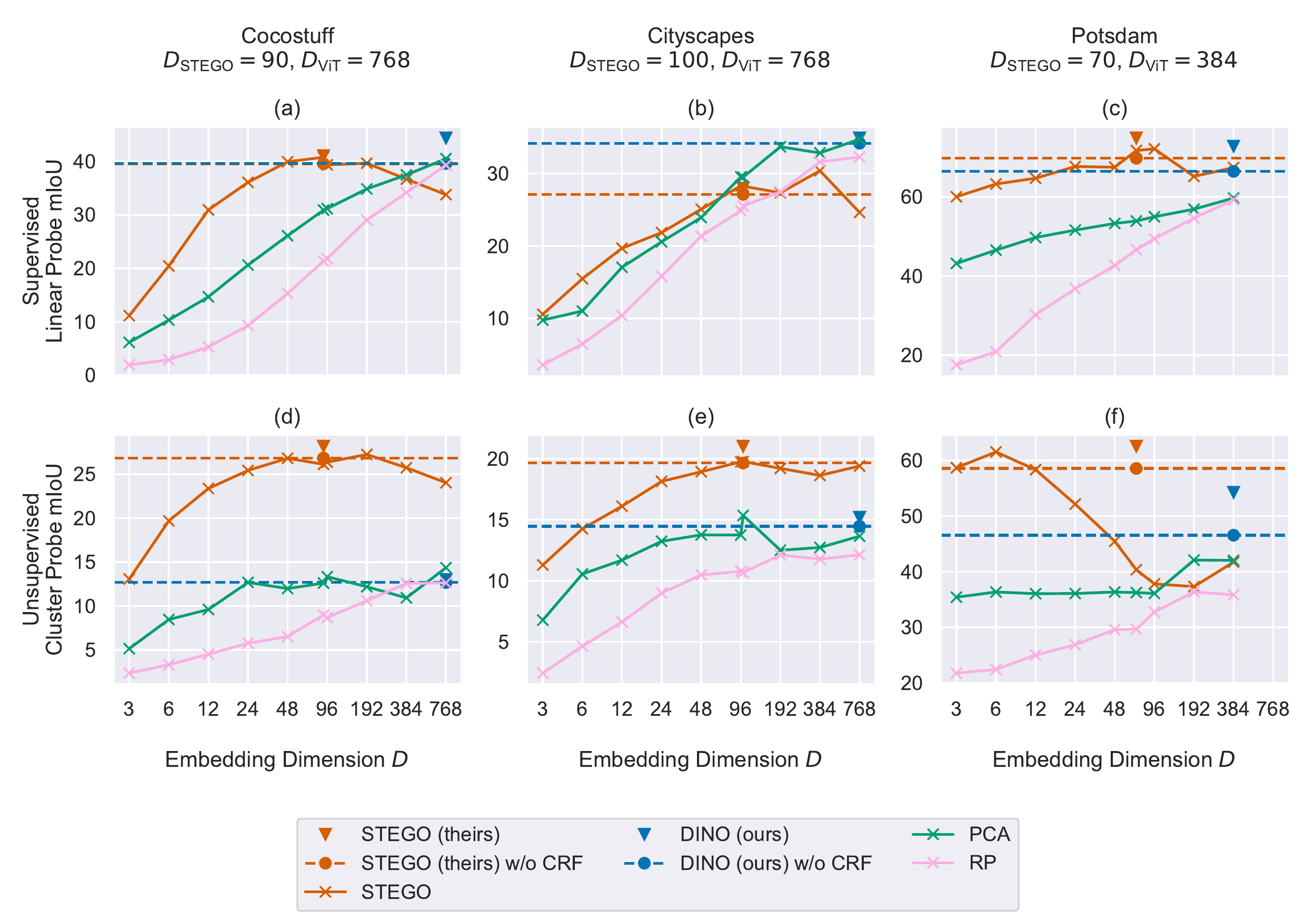}\vspace{-2mm}
    \caption{Validation mIoU of different dimensionality reduction techniques with cluster- and linear probe-style evaluation across different embedding dimensions. The $\blacktriangledown$ data points are the results from \cref{tab:repro_results}. The $\bullet$ data points show the same results without the CRF post-processing. All remaining data is without CRF to paint an undistorted picture of the effectiveness of the underlying segmentation pipeline. The $\times$ data points show the results of models trained with different embedding dimensions and dimensionality reduction algorithms.}
    \label{fig:dims}
    \vspace{-1mm}
\end{figure*}

\cref{fig:dims} shows our results, which analyze the embedding dimension hyperparameter of various unsupervised semantic segmentation techniques. As previously noted, we use three methods to reduce the dimensionality of the DINO features: the segmentation head of STEGO, PCA, and RP. In the latter two cases, we swap out STEGO's segmentation head and train the linear and cluster probes on the PCA- or RP-transformed output directly while keeping the rest of the training and validation pipeline fixed. We observe several interesting results.

For linear probing in \mbox{\cref{fig:dims}~(a--c)}, STEGO can reduce the dimensionality with little loss in performance up to approximately $D_\textrm{STEGO}$. This means that STEGO is an efficient non-linear dimensionality reduction technique. However, through an ablation across multiple dimensions, we confirm our experimental result from \cref{tab:repro_results}, indicating that STEGO provides little, if any, performance benefit over the DINO baseline for linear probe-style evaluation. By analyzing STEGO's performance at the unreduced DINO dimension, \ie, $D_\textrm{STEGO}=D_\textrm{ViT}$, we can isolate the segmentation head's dimensionality reduction from its non-linear projection. Notably, features produced by the STEGO segmentation head at the unreduced dimension are less or equally linearly separable than DINO features. Hence, the segmentation head does not project DINO's features in a useful way for linear probing. However, it should be noted that the hyperparameters of the STEGO training strategy are not fine-tuned and thus are strictly only compatible with the original output dimension $D_\textrm{STEGO}$. Hence, the reduced performance at the full dimension could also be due to hyperparameter instability.
Finally, STEGO performs better or at least on par with the linear dimensionality reduction techniques on the linear probing task.

\mbox{\cref{fig:dims}~(d, e)} shows that STEGO consistently outperforms PCA, RP, and plain DINO across all embedding dimensions in the unsupervised clustering task. This demonstrates that STEGO can robustly project ViT features into a wide range of target dimensions -- even into much lower ones than suggested in the original paper~\cite{hamilton2022unsupervised}. Hence, the segmentation head preserves semantic information for the clustering task while simultaneously reducing the number of embedding dimensions. However, STEGO's superior performance for unsupervised clustering cannot only be attributed to its dimensionality-reducing aspect since we also see a performance boost in~\mbox{\cref{fig:dims}~(d, e)} at the original dimension $D_\textrm{STEGO}=D_\textrm{ViT}=768$. This observation could be evidence of Hamilton \etal's argument of more compact and distinct clusters in features produced by STEGO that could facilitate better $k$-means convergence. However, STEGO's improvement over DINO may also be due to the adaptation of the segmentation head to the new training data distribution. Training the segmentation head or Transformer adapter layers~\cite{houlsbyAdapters} using a simpler loss (\eg, the DINO loss) could potentially achieve comparable performance. Nevertheless, we conclude for the unsupervised downstream task: Besides efficiently reducing the dimension, STEGO also projects DINO's output features to representations that are more suited for $k$-means clustering. In contrast, we did not observe a benefit of STEGO for linear probing over the DINO baseline at an unreduced dimension. 

\mbox{\cref{fig:dims}~(d, e)} also demonstrates that PCA approximately retains the unsupervised performance of DINO up to an embedding dimension of $D=48$ -- a significant reduction by a factor of 16 compared to the original dimension $D_\textrm{ViT}=768$ of the DINO baseline. Simultaneously, in \mbox{\cref{fig:dims}~(a--c)}, the performance of the PCA-transformed features on the linear probing downstream task approximately exponentially decreases with lower embedding dimensions (note that the $x$-axis is in log-scale). 
From \cref{fig:pca}, we know that the first 48 components account for roughly 40\,\% of variance explained for Cityscapes and 25\,\% for Cocostuff. Therefore, lower performance of the PCA-projected features on the linear probing task is expected since the transformed features contain much less information. Surprisingly, the unsupervised clustering performance does not worsen until $D=48$ for Cocostuff and Cityscapes, even though significantly less information is available. Hence, another factor is at play, improving performance for lower embedding dimensions. This factor likely is related to our previous argument that the $k$-means algorithm converges better in lower dimensions. 

Interestingly, all STEGO results in \cref{fig:dims} follow a similar trend, where the cluster and linear performance peaks at a dimension $D_\textrm{STEGO} \ll D_\textrm{ViT}$. Furthermore, \mbox{\cref{fig:dims}~(d, e)} shows a similar trend for the PCA cluster performance, which also has a slight upwards trend at approximately $D_\textrm{STEGO}$. These results suggest some optimal configuration for the $k$-means algorithm, where both effects -- lower dimensions mean less information content, but a better $k$-means convergence -- balance out, again indicating that the dimensionality reduction is a critical contribution to STEGO's performance in the unsupervised clustering case. Another explanation for the peak STEGO performance could be that the hyperparameters, which Hamilton \etal~\cite{hamilton2022unsupervised} identified, work best for $D_\textrm{STEGO}$ and that hyperparameter tuning influences the results at other dimensions.

As shown in \mbox{\cref{fig:dims}~(a, b, d, e)}, we are able to reproduce the originally reported performance of STEGO without CRF for Cocostuff and Cityscapes. However, the Potsdam results in \mbox{\cref{fig:dims}~(f)} behave quite differently than those on the more diverse and larger datasets. While in our previous analysis, we found that all STEGO cluster probe results show a conceptually similar trend (\ie, performance peaks at a dimension $D\ll D_\textrm{ViT}$ and they robustly outperform the baselines), the entire Potsdam curve is shifted to the left and does not meet the expected performance of the pre-trained STEGO model. This vastly different behavior is likely due to two reasons. First, the Potsdam aerial imagery dataset is out of distribution input for the ImageNet-pre-trained DINO backbone, likely leading DINO to produce lower-quality embeddings. Second, the segmentation head is trained on significantly fewer data for Potsdam than Cityscapes and Cocostuff, making it more challenging for the module to compensate for lower-quality DINO features. Note that Potsdam is inherently smaller than Cityscapes and Cocostuff containing only roughly 4.5k training images as opposed to 15k and 490k pre-processed training images for Cityscapes and Cocostuff, respectively.
Nevertheless, it is impressive that STEGO achieves over 60\% mIoU using only 6-dimensional embeddings in \mbox{\cref{fig:dims}~(f)}. A likely reason for much better relative performance at lower dimensions is the considerably easier segmentation task having only 3 classes for Potsdam, as opposed to 27 for Cocostuff and Cityscapes. Separating these three classes using the cluster probe in very high dimensions is challenging, again supporting our argument that STEGO is an efficient dimensionality reduction technique.

Finally, due to computational efficiency, we cache the precomputed DINO features using the Squirrel library~\cite{2022squirrelcore} for PCA and RP. Therefore, we apply a slightly different shuffling than the one used in the case of STEGO. The shuffling styles can influence the performance of the mini-batch $k$-means variant and the linear probe results, which might explain why the PCA and RP results in \cref{fig:dims} sometimes do not match the DINO performance at $D_\textrm{ViT}$ exactly. 
Note that for all STEGO results shown in \cref{fig:dims}, we run the original code with the unaltered shuffling strategy.

Minor results in \cref{fig:dims} include that all other dimensionality reduction techniques in the linear and cluster probe downstream task outperform RP. This is somewhat surprising as RP approximately preserves pair-wise distances. A possible explanation might be the use of a cosine similarity-based $k$-means clustering. Investigating different unsupervised clusterings, such as a standard Euclidian distance-based $k$-means or more advanced clustering approaches, is an interesting direction for future investigations. However, fitting RP is exceptionally fast and compute-efficient. Finally, CRF improves performance by a few percent, as indicated in the STEGO paper~\cite{hamilton2022unsupervised} and other works in the field.

\section{Conclusion}
\label{sec:conclusion}
This work provided a deeper understanding and evaluation of the recent STEGO method for unsupervised semantic segmentation. We uncovered the working mechanisms behind the architecture's segmentation head, highlighted a remarkable performance of the DINO backbone, and found that STEGO compares favorably against traditional dimensionality reduction techniques. A limitation of our analysis is that we did not optimize STEGO's loss parameters for each embedding dimension.

Future research should detail the relationship between the loss parameters and task performance. Additionally, it is essential to investigate whether self-supervised fine-tuning of the DINO backbone on new datasets can yield comparable results to STEGO. Lastly, it is interesting to explore other clustering strategies, such as over-clustering of the features with $k$-means as proposed in MaskDistill~\cite{MaskDistill}, or using different clustering algorithms that perform well in high-dimensional spaces such as \mbox{h-NNE}~\cite{hnne} for which we show first results in the supplementary material.

\vspace{-2mm}
\small{\paragraph{Acknowledgements.} This work was supported by the Federal Ministry for Economic Affairs and Climate Action (BMWK) on the basis of a decision by the German Bundestag as part of the safe.trAIn project under grant no.\ 19I21039G. 
\vspace{-4mm}
\paragraph{Contributions.} All authors designed the study. A.K. implemented and carried out the experiments. A.K. and M.S. wrote the manuscript. M.S. and J.O. supervised the work. 
}

{\small
\bibliographystyle{ieee_fullname}
\bibliography{egbib}

\begin{thebibliography}{10}\itemsep=-1pt

\bibitem{Ahn_2018_CVPR}
Jiwoon Ahn and Suha Kwak.
\newblock Learning pixel-level semantic affinity with image-level supervision
  for weakly supervised semantic segmentation.
\newblock In {\em Conference on Computer Vision and Pattern Recognition}, 2018.

\bibitem{jordan}
Ahmad Alzu'bi and Lujain Alsmadi.
\newblock Monitoring deforestation in {Jordan} using deep semantic segmentation
  with satellite imagery.
\newblock {\em Ecological Informatics}, 2022.

\bibitem{amir2021deep}
Shir Amir, Yossi Gandelsman, Shai Bagon, and Tali Dekel.
\newblock Deep {ViT} features as dense visual descriptors.
\newblock {\em European Conference on Computer Vision Workshop}, 2022.

\bibitem{agri}
Tanmay Anand, Soumendu Sinha, Murari Mandal, Vinay Chamola, and Fei~Richard Yu.
\newblock {AgriSegNet}: Deep aerial semantic segmentation framework for
  {IoT}-assisted precision agriculture.
\newblock {\em IEEE Sensors Journal}, 2021.

\bibitem{binghamRandomProj}
Ella Bingham and Heikki Mannila.
\newblock Random projection in dimensionality reduction: Applications to image
  and text data.
\newblock In {\em International Conference on Knowledge Discovery and Data
  Mining}, 2001.

\bibitem{kMeanswRandomProj}
Christos Boutsidis, Anastasios Zouzias, and Petros Drineas.
\newblock Random projections for $k$-means clustering.
\newblock In {\em Advances in Neural Information Processing Systems}, 2010.

\bibitem{Caesar_2018_CVPR}
Holger Caesar, Jasper Uijlings, and Vittorio Ferrari.
\newblock {COCO-Stuff}: Thing and stuff classes in context.
\newblock In {\em Conference on Computer Vision and Pattern Recognition}, 2018.

\bibitem{swav}
Mathilde Caron, Ishan Misra, Julien Mairal, Priya Goyal, Piotr Bojanowski, and
  Armand Joulin.
\newblock Unsupervised learning of visual features by contrasting cluster
  assignments.
\newblock In {\em Advances in Neural Information Processing Systems}, 2020.

\bibitem{caron2021emerging}
Mathilde Caron, Hugo Touvron, Ishan Misra, Herv\'e J\'egou, Julien Mairal,
  Piotr Bojanowski, and Armand Joulin.
\newblock Emerging properties in self-supervised vision transformers.
\newblock In {\em International Conference on Computer Vision}, 2021.

\bibitem{chen20simclr}
Ting Chen, Simon Kornblith, Mohammad Norouzi, and Geoffrey Hinton.
\newblock A simple framework for contrastive learning of visual
  representations.
\newblock In {\em International Conference on Machine Learning}, 2020.

\bibitem{Cho2021PiCIE}
Jang~Hyun Cho, Utkarsh Mall, Kavita Bala, and Bharath Hariharan.
\newblock {PiCIE}: Unsupervised semantic segmentation using invariance and
  equivariance in clustering.
\newblock In {\em Conference on Computer Vision and Pattern Recognition}, 2021.

\bibitem{Collins_2018_ECCV}
Edo Collins, Radhakrishna Achanta, and Sabine Susstrunk.
\newblock Deep feature factorization for concept discovery.
\newblock In {\em European Conference on Computer Vision}, 2018.

\bibitem{Cordts_2016_CVPR}
Marius Cordts, Mohamed Omran, Sebastian Ramos, Timo Rehfeld, Markus Enzweiler,
  Rodrigo Benenson, Uwe Franke, Stefan Roth, and Bernt Schiele.
\newblock The {Cityscapes} dataset for semantic urban scene understanding.
\newblock In {\em Conference on Computer Vision and Pattern Recognition}, 2016.

\bibitem{DasguptaRandomProj}
Sanjoy Dasgupta.
\newblock Experiments with random projection.
\newblock In {\em Conference on Uncertainty in Artificial Intelligence}, 2000.

\bibitem{mnist}
Li Deng.
\newblock The {MNIST} database of handwritten digit images for machine learning
  research {[Best of the Web]}.
\newblock {\em IEEE Signal Processing Magazine}, 2012.

\bibitem{dosovitskiy2021an}
Alexey Dosovitskiy, Lucas Beyer, Alexander Kolesnikov, Dirk Weissenborn,
  Xiaohua Zhai, Thomas Unterthiner, Mostafa Dehghani, Matthias Minderer, Georg
  Heigold, Sylvain Gelly, Jakob Uszkoreit, and Neil Houlsby.
\newblock An image is worth 16$\times$16 words: Transformers for image
  recognition at scale.
\newblock In {\em International Conference on Learning Representations}, 2021.

\bibitem{relu}
Xavier Glorot, Antoine Bordes, and Yoshua Bengio.
\newblock Deep sparse rectifier neural networks.
\newblock In {\em International Conference on Artificial Intelligence and
  Statistics}, 2011.

\bibitem{byol}
Jean-Bastien Grill, Florian Strub, Florent Altché, Corentin Tallec, Pierre~H.
  Richemond, Elena Buchatskaya, Carl Doersch, Bernardo~Avila Pires,
  Zhaohan~Daniel Guo, Mohammad~Gheshlaghi Azar, Bilal Piot, Koray Kavukcuoglu,
  Rémi Munos, and Michal Valko.
\newblock Bootstrap your own latent -- a new approach to self-supervised
  learning.
\newblock In {\em Advances in Neural Information Processing Systems}, 2020.

\bibitem{ganav}
Tianrui Guan, Divya Kothandaraman, Rohan Chandra, Adarsh~Jagan Sathyamoorthy,
  Kasun Weerakoon, and Dinesh Manocha.
\newblock {GA-Nav}: Efficient terrain segmentation for robot navigation in
  unstructured outdoor environments.
\newblock {\em IEEE Robotics and Automation Letters}, 2022.

\bibitem{hamilton2022unsupervised}
Mark Hamilton, Zhoutong Zhang, Bharath Hariharan, Noah Snavely, and William~T.
  Freeman.
\newblock Unsupervised semantic segmentation by distilling feature
  correspondences.
\newblock In {\em International Conference on Learning Representations}, 2022.

\bibitem{houlsbyAdapters}
Neil Houlsby, Andrei Giurgiu, Stanislaw Jastrzebski, Bruna Morrone, Quentin
  De~Laroussilhe, Andrea Gesmundo, Mona Attariyan, and Sylvain Gelly.
\newblock Parameter-efficient transfer learning for {NLP}.
\newblock In {\em International Conference on Machine Learning}, 2019.

\bibitem{Ji_2019_ICCV}
Xu Ji, Joao~F. Henriques, and Andrea Vedaldi.
\newblock Invariant information clustering for unsupervised image
  classification and segmentation.
\newblock In {\em International Conference on Computer Vision}, 2019.

\bibitem{Lindenstrauss1984}
William Johnson and Joram Lindenstrauss.
\newblock Extensions of {Lipschitz} maps into a {Hilbert} space.
\newblock {\em Contemporary Mathematics}, 1984.

\bibitem{crf}
Philipp Kr\"{a}henb\"{u}hl and Vladlen Koltun.
\newblock Efficient inference in fully connected {CRFs} with {Gaussian} edge
  potentials.
\newblock In {\em Advances in Neural Information Processing Systems}, 2011.

\bibitem{cifar}
Alex Krizhevsky and Geoffrey Hinton.
\newblock Learning multiple layers of features from tiny images.
\newblock {\em Technical Report, University of Toronto}, 2009.

\bibitem{kuhn1955hungarian}
Harold~W Kuhn.
\newblock The {Hungarian} method for the assignment problem.
\newblock {\em Naval Research Logistics Quarterly}, 1955.

\bibitem{Lai_2021_CVPR}
Xin Lai, Zhuotao Tian, Li Jiang, Shu Liu, Hengshuang Zhao, Liwei Wang, and
  Jiaya Jia.
\newblock Semi-supervised semantic segmentation with directional context-aware
  consistency.
\newblock In {\em Conference on Computer Vision and Pattern Recognition}, 2021.

\bibitem{MacQueen1967}
J.~B. MacQueen.
\newblock Some methods for classification and analysis of multivariate
  observations.
\newblock In {\em Berkeley Symposium on Mathematical Statistics and
  Probability}, 1967.

\bibitem{Melas-Kyriazi_2022_CVPR}
Luke Melas-Kyriazi, Christian Rupprecht, Iro Laina, and Andrea Vedaldi.
\newblock Deep spectral methods: A surprisingly strong baseline for
  unsupervised semantic segmentation and localization.
\newblock In {\em Conference on Computer Vision and Pattern Recognition}, 2022.

\bibitem{Oh_2021_CVPR}
Youngmin Oh, Beomjun Kim, and Bumsub Ham.
\newblock Background-aware pooling and noise-aware loss for weakly-supervised
  semantic segmentation.
\newblock In {\em Conference on Computer Vision and Pattern Recognition}, 2021.

\bibitem{pca}
Karl Pearson.
\newblock On lines and planes of closest fit to systems of points in space.
\newblock {\em The London, Edinburgh, and Dublin Philosophical Magazine and
  Journal of Science}, 1901.

\bibitem{unet}
Olaf Ronneberger, Philipp Fischer, and Thomas Brox.
\newblock {U-Net}: Convolutional networks for biomedical image segmentation.
\newblock In {\em Medical Image Computing and Computer-Assisted Intervention},
  2015.

\bibitem{imagenet2015}
Olga Russakovsky, Jia Deng, Hao Su, Jonathan Krause, Sanjeev Satheesh, Sean Ma,
  Zhiheng Huang, Andrej Karpathy, Aditya Khosla, Michael Bernstein,
  Alexander~C. Berg, and Li Fei-Fei.
\newblock {ImageNet Large Scale Visual Recognition Challenge}.
\newblock {\em International Journal of Computer Vision}, 2015.

\bibitem{her2net}
Monjoy Saha and Chandan Chakraborty.
\newblock {Her2Net}: A deep framework for semantic segmentation and
  classification of cell membranes and nuclei in breast cancer evaluation.
\newblock {\em IEEE Transactions on Image Processing}, 2018.

\bibitem{hnne}
M.~Saquib Sarfraz, Marios Koulakis, Constantin Seibold, and Rainer
  Stiefelhagen.
\newblock Hierarchical nearest neighbor graph embedding for efficient
  dimensionality reduction.
\newblock In {\em Conference on Computer Vision and Pattern Recognition}, 2022.

\bibitem{surveyLabelEfficient}
Wei Shen, Zelin Peng, Xuehui Wang, Huayu Wang, Jiazhong Cen, Dongsheng Jiang,
  Lingxi Xie, Xiaokang Yang, and Q. Tian.
\newblock A survey on label-efficient deep image segmentation: Bridging the gap
  between weak supervision and dense prediction.
\newblock {\em IEEE Transactions on Pattern Analysis and Machine Intelligence},
  2023.

\bibitem{normcuts}
Jianbo Shi and Jitendra Malik.
\newblock Normalized cuts and image segmentation.
\newblock {\em IEEE Transactions on Pattern Analysis and Machine Intelligence},
  2000.

\bibitem{2022squirrelcore}
Squirrel~Developer Team.
\newblock Squirrel: A {Python} library that enables {ML} teams to share, load,
  and transform data in a collaborative, flexible, and efficient way.
\newblock {\em GitHub. Note:
  https://github.com/merantix-momentum/squirrel-core}, 2022.

\bibitem{tschernezki22neural}
Vadim Tschernezki, Iro Laina, Diane Larlus, and Andrea Vedaldi.
\newblock Neural feature fusion fields: {3D} distillation of self-supervised
  {2D} image representations.
\newblock In {\em International Conference on {3D} Vision}, 2022.

\bibitem{MaskContrast}
Wouter Van~Gansbeke, Simon Vandenhende, Stamatios Georgoulis, and Luc Van~Gool.
\newblock Unsupervised semantic segmentation by contrasting object mask
  proposals.
\newblock In {\em International Conference on Computer Vision}, 2021.

\bibitem{MaskDistill}
Wouter Van~Gansbeke, Simon Vandenhende, and Luc Van~Gool.
\newblock Discovering object masks with transformers for unsupervised semantic
  segmentation.
\newblock In {\em arXiv}, 2022.

\bibitem{Wang_2020_CVPR}
Yude Wang, Jie Zhang, Meina Kan, Shiguang Shan, and Xilin Chen.
\newblock Self-supervised equivariant attention mechanism for weakly supervised
  semantic segmentation.
\newblock In {\em Conference on Computer Vision and Pattern Recognition}, 2020.

\bibitem{setr}
Sixiao Zheng, Jiachen Lu, Hengshuang Zhao, Xiatian Zhu, Zekun Luo, Yabiao Wang,
  Yanwei Fu, Jianfeng Feng, Tao Xiang, Philip~H.S. Torr, and Li Zhang.
\newblock Rethinking semantic segmentation from a sequence-to-sequence
  perspective with transformers.
\newblock In {\em Conference on Computer Vision and Pattern Recognition}, 2021.

\end{thebibliography}


\begin{thebibliography}{1}\itemsep=-1pt

\bibitem{hamilton2022unsupervised}
Mark Hamilton, Zhoutong Zhang, Bharath Hariharan, Noah Snavely, and William~T.
  Freeman.
\newblock Unsupervised semantic segmentation by distilling feature
  correspondences.
\newblock In {\em International Conference on Learning Representations}, 2022.

\bibitem{adam2015}
Diederik~P. Kingma and Jimmy Ba.
\newblock Adam: A method for stochastic optimization.
\newblock In {\em International Conference on Learning Representations}, 2015.

\bibitem{kuhn1955hungarian}
Harold~W Kuhn.
\newblock The {Hungarian} method for the assignment problem.
\newblock {\em Naval Research Logistics Quarterly}, 1955.

\bibitem{McInnes2018}
Leland McInnes, John Healy, Nathaniel Saul, and Lukas Großberger.
\newblock {UMAP}: Uniform manifold approximation and projection.
\newblock {\em Journal of Open Source Software}, 2018.

\bibitem{hnne}
M.~Saquib Sarfraz, Marios Koulakis, Constantin Seibold, and Rainer
  Stiefelhagen.
\newblock Hierarchical nearest neighbor graph embedding for efficient
  dimensionality reduction.
\newblock In {\em Conference on Computer Vision and Pattern Recognition}, 2022.

\bibitem{tsne}
Laurens van~der Maaten and Geoffrey Hinton.
\newblock Visualizing data using {t-SNE}.
\newblock {\em Journal of Machine Learning Research}, 2008.

\end{thebibliography}
}

\end{document}


\title{Supplementary Material: Uncovering the Inner Workings of STEGO for \\Safe Unsupervised Semantic Segmentation}


\author{
Alexander Koenig 
\qquad
Maximilian Schambach
\qquad
Johannes Otterbach
\\\\
Merantix Momentum\\
{\tt\small \{firstname.lastname\}@merantix.com}
}

\maketitle

\section{Full Configuration}
We showed the model configuration in detail in our main contribution. 
We noticed that the original STEGO configuration varies for the three datasets. 
\cref{tab:config_all} shows some remaining noteworthy parameters in STEGO's configuration, which are the same for all datasets.

\section{Accuracy Results}
\begin{figure*}[t]
    \centering
    \includegraphics[width=\linewidth,trim={0 2.22cm 0 0},clip]{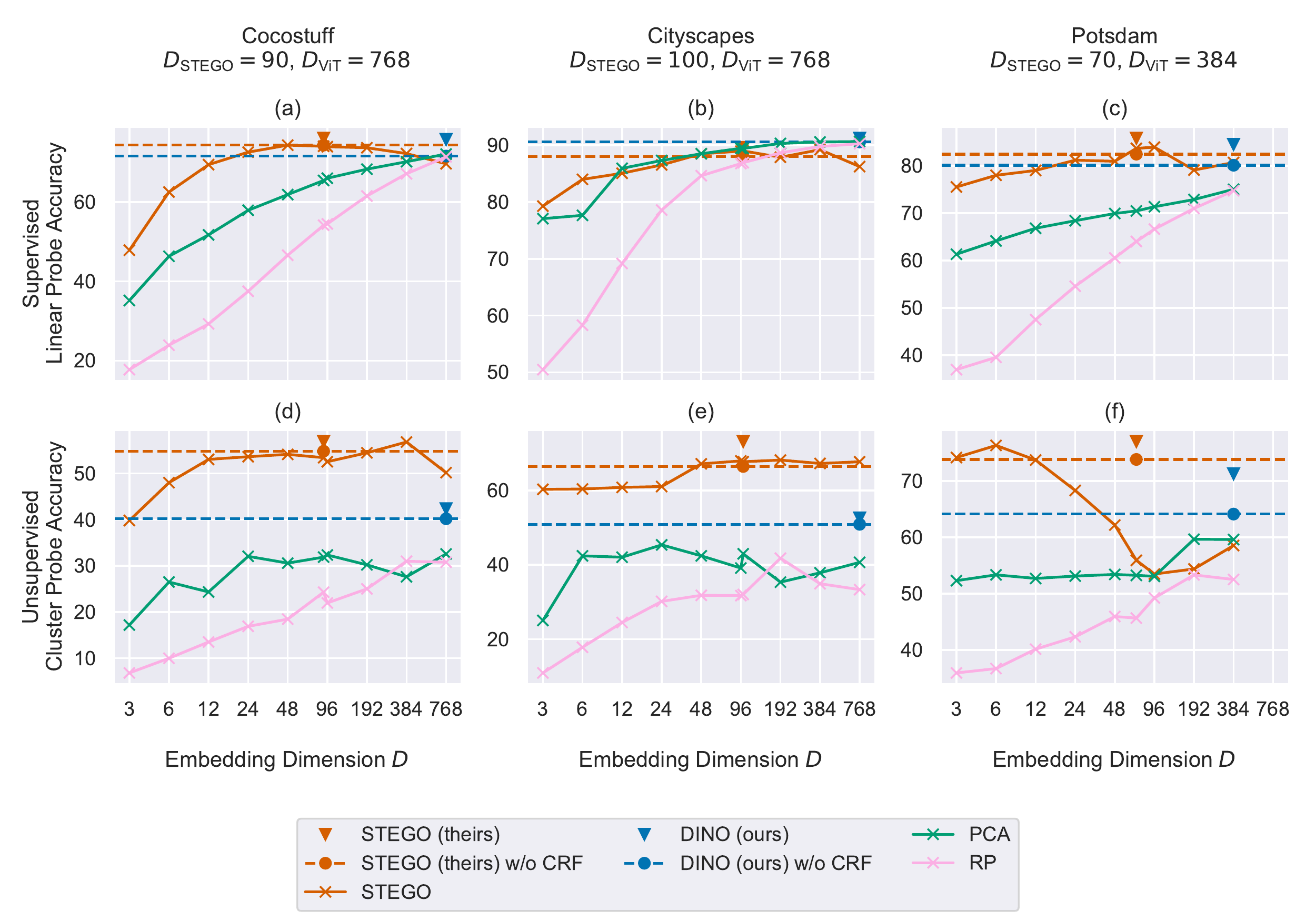}
    \caption{Validation accuracy of different dimensionality reduction techniques. Readers are referred to \cref{fig:hnne} for a color-coded legend.}
    \label{fig:dims_acc}
\end{figure*}

In the main paper, we report the mIoU results of the different frameworks. Here, we also provide supplementary information on the accuracy metric in \cref{fig:dims_acc}. Our observations from our previous discussion also translate to the accuracy plots. Hence, we solely display the results for completeness without further discussion.

\section{Non-linear Dimension Reduction Baseline}
Our previous analysis compared STEGO with the linear dimensionality reduction techniques, PCA, RP, and the DINO baseline. Since STEGO is a non-linear projection, comparing it to a non-linear dimension reduction method is interesting. Initially, we investigated Uniform Manifold Approximation and Projection (UMAP) \cite{McInnes2018}, which builds a fuzzy topological representation of the data in the original space and, via cross-entropy, searches a new representation that approximates this topology in a lower dimensional space. Despite improved scalability of the UMAP algorithm over other non-linear dimension reduction algorithms like t-SNE \cite{tsne}, UMAP was prohibitively expensive to compute across all datasets and different embedding dimensions (\eg, there are $\approx$\,0.8 billion 768-dimensional ViT training tokens for the Cocostuff dataset alone). A recently proposed optimization-free and faster algorithm, Hierarchical Nearest Neighbor Embedding (h-NNE) \cite{hnne}, approaches the problem by first building a clustering hierarchy of the data in high dimensions. Afterward, the method hierarchically projects the data into a lower dimensional space, preserving 1-nearest neighbor relationships.

\begin{table}[t]
\centering
\small
\begin{tabular}{lrrrr}
\toprule

Parameter & Value \\ 
\midrule
{Loader crop type} & {Center}  \\ 
{Extra clusters} & {0}  \\ 
{Optimizers} & {Adam \cite{adam2015}}  \\ 
{Linear and cluster probe learning rates} & {0.005}  \\ 
{Segm. head learning rate} & {0.0005}  \\ 
{Segm. head dropout probability} & {0.1}  \\ 
{Feature samples} & {11}  \\ 
{Negative samples} & {5}  \\
\bottomrule
\end{tabular}
\caption{Remaining model configuration for STEGO. These are the original parameters from the paper, also used in our study. Only the last four parameters are specific to the training of the segmentation head -- the others also apply to our DINO, PCA, and RP baselines. Hamilton \etal's \cite{hamilton2022unsupervised} code repository contains more information on the parameters.}
\label{tab:config_all}
\vspace{-0.5cm}
\end{table}

\begin{figure*}[t]
    \centering
    \includegraphics[width=\linewidth]{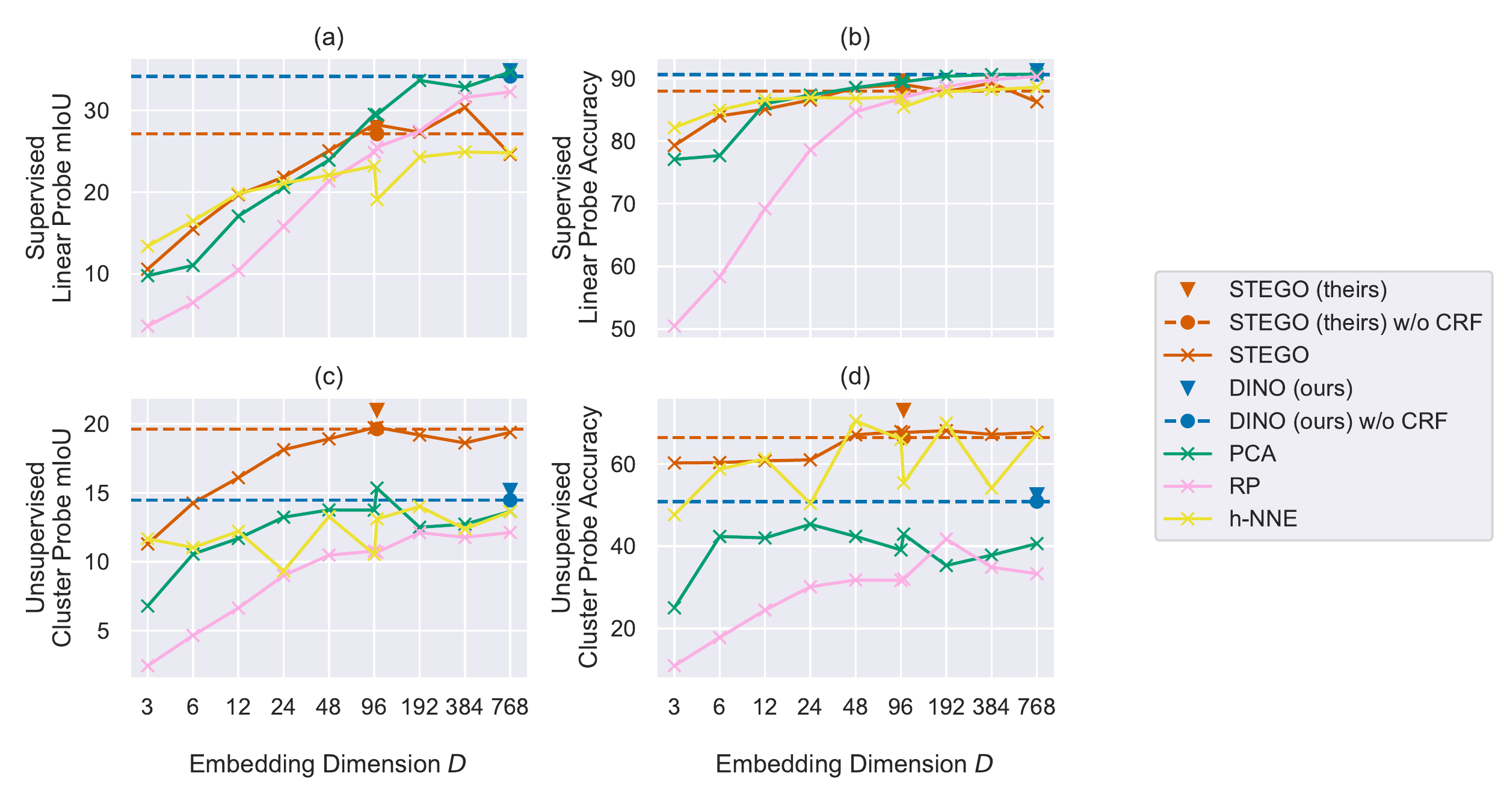}
    \caption{Cityscapes validation results from our main contribution and \cref{fig:dims_acc} with overlaid h-NNE \cite{hnne} results.}
    \label{fig:hnne}
\end{figure*}

We fit h-NNE on a randomly sampled subset of 1 million ViT training tokens, project the entire training and validation set into lower dimensions, and fit the linear and cluster probes on these projected embeddings. \cref{fig:hnne} shows the results for the Cityscapes dataset. In \mbox{\cref{fig:hnne}~(a,b)}, the h-NNE algorithm shows similar performance on the linear probing downstream task as STEGO, PCA, and RP across a wide range of dimensions. For the unsupervised cluster probe in \mbox{\cref{fig:hnne}~(c)}, we see approximately equal mIoU performance compared to the PCA baseline, while the accuracy of the cluster probe trained on the h-NNE projections in \mbox{\cref{fig:hnne}~(d)} outperforms the PCA, RP baselines, although the variance of the h-NNE results appears more significant. 

In summary, these preliminary results show that the non-linear projection with the h-NNE algorithm yields little to no benefit over the linear projection methods in the tested benchmarks. However, we assume that clustering the h-NNE projected output with $k$-means might not be the most suitable unsupervised downstream evaluation. The h-NNE algorithm already provides a hierarchy of clusterings. Hence, in future work, one could directly map the clusters detected by h-NNE to the human-interpretable labels with the Hungarian algorithm \cite{kuhn1955hungarian}.

{\small
\bibliographystyle{ieee_fullname}
\bibliography{egbib}
}